\documentclass[runningheads]{llncs}
\usepackage[T1]{fontenc}
\usepackage{graphicx}
\usepackage{tabularx}
\usepackage{todonotes}
\usepackage{amsmath}

\usepackage{hyperref}

\begin{document}
\title{HyDI: A hybrid Deep Learning-Inductive Logic Programming ensemble for multi-label classification}
\titlerunning{HyDI}
%
\author{Simon Flügel\inst{1}\orcidID{0000-0003-3754-9016} \and
Till Mossakowski\inst{1}\orcidID{0000-0002-8938-5204} 
}
\authorrunning{S. Flügel et al.}
%
\institute{Institute for Computer Science, University of Osnabr\"uck, Neuer Graben 29, 49074 Osnabr\"uck, Germany 
\email{sfluegel@uni-osnabrueck.de}\\
}
\maketitle              
\begin{abstract}
While attaining remarkable results for many applications, Deep Learning models are notoriously difficult to explain.
This work introduces HyDI, a hybrid ensemble architecture for hierarchical multi-label classification. It combines a Deep Learning (DL) model with rule-based classifiers generated by Inductive Logic Programming (ILP). 
For leaf classes of the label hierarchy, the rule-based classifiers replace the DL model, leading to more transparent classification results.
HyDI is applied to the Chemical Entities of Biological Interest (ChEBI) ontology, providing ILP-generated rules for 314 classes.
For these classes, HyDI can generate global explanations as well as local explanations that combine visual and text-based descriptions.

\end{abstract}
\section{Introduction}
Since the advent of the attention mechanism, Deep Learning (DL) methods have been at the forefront of Machine Learning research, with success in a wide range of applications such as image classification, text generation or bioinformatics.
Nevertheless, DL models on their own are not a cure-all. 
Their results are inherently unexplainable, they rely on huge amounts of data and are expensive to train. 
Neuro-symbolic integration is trying to fill this gap by combining neural approaches with symbolic ones. 

This work in particular addresses the task of hierarchical multi-label classification (HMLC) in which an object may have several labels, which themselves form a hierarchy. 
A major challenge in HMLC is class imbalance. While classes higher up in the hierarchy have a high number of associated objects, classes further down become increasingly rare. 

The \textbf{Hy}brid \textbf{D}eep Learning-\textbf{I}nductive Logic Programming (HyDI) ensemble is a neuro-symbolic ensemble architecture for HMLC. 
It has three components: (1) A Deep Learning model as a native multi-label classifier that makes predictions for all classes simultaneously. (2) Symbolic Binary Relevance classifiers that make decisions for individual leaf classes, generated with Inductive Logic Programming (ILP).
(3) An aggregation mechanism that decides which model to use for which classes.
The goal of the ensemble is to maintain the classification performance of the DL model while making the results more explainable. Since the leaf classes are particularly important to the overall classification, HyDI introduces symbolic classifiers for this part of the label hierarchy.

In a case study on the Chemical Entities of Biological Interest (ChEBI) ontology~\cite{malik2026chebi}, HyDI is applied to a real-world HMLC task. 
To improve the understandability of ILP rules for domain experts, HyDI features global and local explanations that combine natural language texts with visualisations.

The remainder of this paper is structured as follows. Section~\ref{sec:related-work} covers Related Work. In Section~\ref{sec:methodology} the HyDI ensemble and the ChEBI case study are introduced. Section \ref{sec:results} evaluates the learned ILP rules and the ensemble performance on ChEBI. Conclusions and Future Work are discussed in Section~\ref{sec:conclusions}.

\section{Related work}\label{sec:related-work}

\subsection{ILP explainability}
Explainability is a commonly cited strength of ILP approaches~\cite{zhang2023critical}. Here, we focus on approaches that combine a DL model with ILP to enhance explainability.

\cite{raboldExplainingBlackBoxClassifiers2018} generate local explanations for an image classification task. The approach assumes a black-box classifier (here, a Convolution Neural Network, CNN). Adapting the LIME~\cite{ribeiro2016should} algorithm, the classifier is fed modified inputs designed to identify which parts of the input image were relevant for the prediction. The final explanations are generated by the ILP system Aleph~\cite{srinivasan2001aleph},
While \cite{raboldExplainingBlackBoxClassifiers2018} relies on pre-defined visual attributes, \cite{raboldEnrichingVisualVerbal2020} extends this by automatic attribute extraction.

A similar methodology is proposed by \cite{dasaroInductiveLogicProgramming2020}, also adapting LIME for local explanations. Here, the ILP system ILASP~\cite{law2020ilasp} is used to explain Support Vector Machine (SVM) predictions on preference learning tasks. In addition, \cite{dasaroInductiveLogicProgramming2020} also produces global explanations (i.e., explanations for a whole class instead of individual samples). This is done by training ILASP on the SVM predictions, aiming for an approximation of the SVM.

\cite{raboldExpressiveExplanationsDNNs2020} focuses on the generation of global explanations as well. For image data, concept embedding analysis is applied to the intermediate output of a CNN to extract semantic concepts. These concepts are used as background knowledge for an ILP system. The goal of the ILP system is to explain the CNN's predictions based on the extracted concepts.

This work differs from these approaches in that we do not use the ILP system to explain the DL predictions post-hoc. Instead, 
ILP rules are learned directly from the input data. 
While this would be infeasible in the image classification domain, structured domains like organic chemistry are more favourable to this approach~\cite{srinivasan1996theories}. 
This comes at the cost that we cannot make deductions about the inner workings of the DL model. 
Instead, by integrating the DL model into an ensemble, we can provide direct explanations for classes that are predicted by an ILP rule, bypassing the DL model. For our use case, this type of explanation is more valuable.

A similarity to \cite{raboldEnrichingVisualVerbal2020} in particular is the connection of ILP rules to visual explanations. \cite{raboldEnrichingVisualVerbal2020} highlights image regions that were needed to satisfy an ILP rule. 
We apply a similar technique to chemical compounds, highlighting atoms and relating them to the explanation. However, while \cite{raboldEnrichingVisualVerbal2020} only does this as a proof of concept, we provide visual explanations as part of an automated pipeline.

\subsection{Multi-label classification for ChEBI}
In the context of ChEBI, both DL-based and symbolic classification approaches have been proposed so far. 

In \cite{glauer2024interpretable}, Transformer models based on SMILES strings, a textual representation of molecules, are used to classify molecule into ChEBI classes. 
Local explanations for the Transformer predictions are generated from the attention relation between different parts of the input. Connecting the attentions to the molecule graph, this allows for a visual insight into which parts of the molecule were likely relevant for classification. 

A different direction is taken by \cite{memariani2025box} which also uses a Transformer model to encode SMILES strings into a latent representation. However, as an additional component,  each label gets learned as a box in the latent space (i.e., pairs of coordinates). 
This allows for an intuitive interpretation of classes and can be used to identify disjointness or subsumption relations between classes.

While these approaches allow for some insights into the model, they don't provide a full explanation of a classification. 
Here, symbolic approaches have the advantage of being inherently traceable. 
For instance, \cite{magkaRulebasedOntologicalFramework2014} models 51 chemical classes with logic programs. Alongside the programs, they provide a classification tool that automatically classifies molecules according to the programs. To improve the accessibility of their rules, \cite{magkaRulebasedOntologicalFramework2014} propose a surface syntax that is close to natural language.

\cite{flugel2026defining} formalises 14 ChEBI classes (as well as 53 non-ChEBI classes) related to peptides in monadic second-order logic (MSOL). 
These definitions are subsequently used in a classification tool which translates the MSOL definitions into a performant Python implementation~\cite{flugel2025chemlog}. 
For these definitions, local explanations are generated via model checking. 
The functional groups leading to the classification are identified and highlighted in the molecule graph.

The main drawback of these approaches is that the underlying rules have to be constructed manually. 
The goal of this work is to automate the rule construction process, generating logical rules directly from the input data with ILP. Also, we aim to produce local explanations for the ILP-generated rules, similar to the style of \cite{flugel2026defining}.

A strategy to automate the generation of symbolic classifiers has been proposed by \cite{mungall2025chemical}. 
Their Chemical Classifier Programs (CCPs) are LLM-generated Python programs, prompted with the natural language definitions present in ChEBI. 
To improve interpretability, the CCPs return a natural language explanation alongside the predictions.
In contrast to \cite{mungall2025chemical}, our work taps into a different source of information (molecule samples instead of natural language definitions). 

In \cite{flugelHeterogeneousEnsembleLearning2026}, the logical definitions from \cite{flugel2026defining} are combined with DL models in an ensemble learning approach, called Chebifier 2. 
For each class, the logical definitions and several DL models are aggregated, weighting each model according to its performance. 
The HyDI ensemble proposed in this work differs from Chebifier 2 not only in the inclusion of ILP models, but also in its architecture: Instead of weighting several predictions, each label gets predicted by one classifier (either the DL model or an ILP rule). This allows for clearer explanations: If several predictions are weighted, the final result cannot be attributed to a single model.

\section{Methodology}
\label{sec:methodology}
In this section, we first give a short introduction to ILP (Section~\ref{sec:meth-ilp}), followed by an explanation of the HyDI ensemble architecture. In Section~\ref{sec:meth-chebi}, the HyDI ensemble is applied to the ChEBI ontology. The global and local explanations are introduced in Sections~\ref{sec:meth-global} and~\ref{sec:meth-local}. Section~\ref{sec:meth-exp} discusses the experimental configuration used during the evaluation.

\subsection{ILP}\label{sec:meth-ilp}
Inductive Logic Programming (ILP)~\cite{muggleton1991inductive,cropperInductiveLogicProgramming2022} is a machine learning methodology in which a hypothesis is learned from background knowledge (BK), represented as logic programs. Given positive and negative examples, BK and a target predicate, the ILP system's goal is to find a hypothesis for the target predicate that classifies the examples correctly.
A wide range of ILP systems exist which differ in how the hypothesis space is searched~\cite{cropperInductiveLogicProgramming2022}. 

This work uses the ILP system Popper~\cite{cropperLearningProgramsLearning2021} since it supports noise~\cite{cropperInductiveLogicProgramming2022} and has been shown to perform well on large domains~\cite{cropperLearningLogicPrograms2023}.
Popper is a meta-level ILP system implementing the ``learning from failures'' approach, in which the hypothesis space is iteratively restricted based on misclassifications. In each iteration, Popper generates a new hypothesis. This hypothesis is tested against examples and BK with Prolog. Based on misclassified examples, new Answer Set Programming (ASP) constraints are added to guide hypothesis generation.
%
The hypothesis $h$ selected by Popper optimises the minimal description length (MDL) cost function
\begin{equation}
    cost(h) = size(h) + \mathit{FN}(h) + \mathit{FP}(h)
\end{equation}
where $size$ is the number of literals, $\mathit{FN}$ the number of false negatives (FNs) and $\mathit{FP}$ the number of false positives (FPs) for a given set of examples and BK.
In this work, we extend this cost function by introducing a bias towards either false negatives or false positives:
\begin{equation}
    cost(h) = size(h) + \alpha \mathit{FN}(h) + \beta \mathit{FP}(h)
\end{equation}
In the evaluation, 3 settings will be used, denoted as \textbf{ILP-FP} ($\alpha=1,\beta=3$), \textbf{ILP-FN} ($\alpha=3,\beta=1$) and \textbf{ILP-REG} ($\alpha=1,\beta=1$), which corresponds to the original MDL cost function.
The motivation for this is to generate a broader range of hypotheses for each ILP problem out of which the ensemble can select the best one. 

\subsection{The HyDI ensemble}
\label{sec:meth-hydi}
\begin{figure}[tb]
    \centering
    \includegraphics[width=0.5\linewidth]{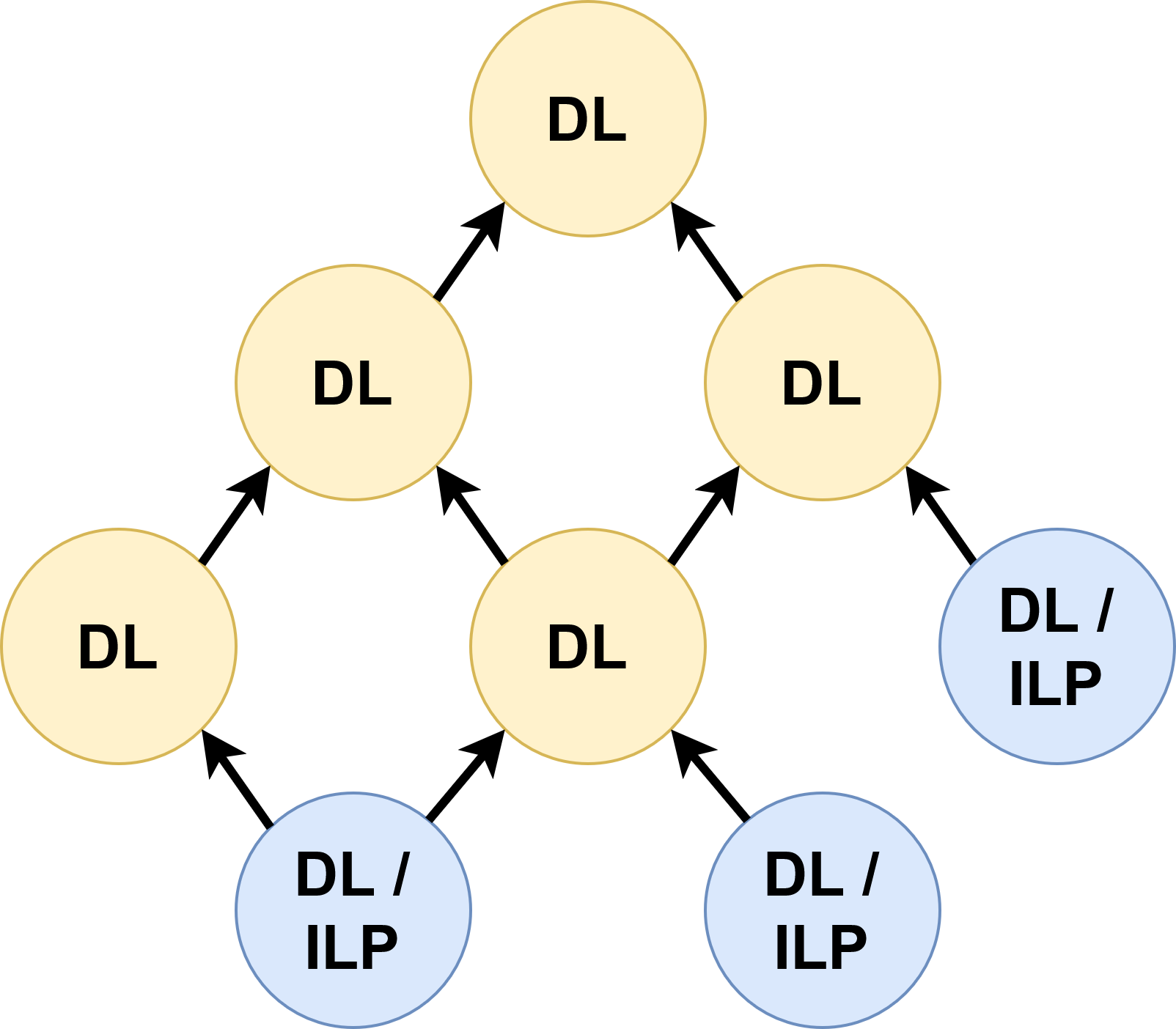}
    \caption{Abstract visualisation of how HyDI is applied to a HMLC problem. Non-leaf classes (highlighted in yellow) are predicted by a DL model. Leaf classes (highlighted in blue) are predicted by either a DL model or an ILP rule (depending on performance).}
    \label{fig:hyDI-architecture}
\end{figure}

Figure~\ref{fig:hyDI-architecture} shows the architecture of the HyDI ensemble. 
The classes of the HMLC problem are divided into 2 groups: Leaf classes and non-leaf classes. Non-leaf classes are predicted by a native multi-label classifier, the DL model. 
The motivation for this is that for one, classes higher up in the hierarchy have more available samples, making them more suitable for DL. The other motivation is the hierarchy depth. An alternative to using a native multi-label classifier would be a binary relevance approach in which a binary classifier is trained for each class independently (in our case, this could be either a DL model or an ILP rule). 
However, this would multiply the number of potential failure points. For example, assume that all binary classifiers have a recall of $0.95$ and are statistically independent. A class at depth 10 in the hierarchy would only have a recall of $0.95^{10}\approx0.60$ since a sample would have to be classified correctly by 10 successive classifiers. This issue gets worse when considering multiple inheritance. Since a class can have several superclasses, it relies on \textit{all} superclasses (and their superclasses in turn) to be predicted correctly.

For leaf nodes, the ensemble has to choose between the DL prediction and one of the ILP predictions. 
This decision is based on the validation F1 score of class $c$: 
\begin{equation}
    \mathit{F1}(c) = \frac{2 \mathit{TP}(c)}{2 \mathit{TP}(c) + \mathit{FP}(c) + \mathit{FN}(c)}
    \label{eq:f1-score}
\end{equation}
For the DL model, this can be calculated directly from the true positives (TPs), false positives (FPs) and false negatives (FNs) on the validation set.
The ILP rules however are designed as binary classifiers, meaning that each rule decides whether a sample belongs to $c$ under the condition that the sample belongs to the superclasses of $c$. 
Therefore, the classification performance depends on the performance of the DL model for the superclasses of $c$:
For ILP rules, TPs are samples that have a positive label, are classified correctly by the ILP rule for class $c$ and are classified correctly by the DL model for all direct superclasses of $c$. FNs are samples that have a positive label but are classified incorrectly by the ILP rule for $c$ or by the DL model for at least one of the direct superclasses. FPs are samples with a negative label that are misclassified by the ILP rule for $c$ as well as the DL model for all direct superclasses.

With this method, the validation F1 score is calculated per class for all ILP rules and the DL model. The model or rule with the highest F1 score gets chosen for the ensemble. 
In case of ties, ILP rules get preference over the DL model. If several ILP rules have the same F1 score, the mean of the ILP configuration's F1 scores across all produced rules (the macro F1 score) is used as a tie breaker.

\subsection{Application to ChEBI}
\label{sec:meth-chebi}

\subsubsection{ChEBI25-3STAR dataset}
The Chemical Entities of Biological Interest (ChEBI) ontology (version 248) consists of 205,132 classes, 62,002 of which are manually curated (the 3-STAR subset).
ChEBI classes are structured hierarchically, starting with broad concepts such as \textit{organic molecular entity} (CHEBI:50860) and moving on to more specific concepts (e.g. \textit{primary alcohol}, CHEBI:15734) and individual molecules (e.g. \textit{methanol}, CHEBI:17790).
The goal of the classification task is to assign all superclasses (including indirect ones) to molecules. For instance, given the molecule \textit{methanol}, a model would have to identify \textit{primary alcohol} and \textit{organic molecular entity} (among others) as superclasses.

To use ChEBI for this classification task, we need to introduce the concepts of \textbf{label classes}, \textbf{leaf classes} and \textbf{molecule classes}.
Following the approach developed by~\cite{glauer2024chebifier}, all classes that are annotated with a molecule structure are molecule classes. Label classes are defined by having at least a given number of subclasses that are molecule classes (here, 25). 
Leaf classes are label classes that have no subclasses that are also label classes.
Contrary to~\cite{glauer2024chebifier}, this work only uses molecule classes from the 3-STAR subset. This excludes classes that were added by automated tools and ensures that models only train on human-curated data.
This results in the ChEBI25-3STAR dataset with a total of 1,808 labels (out of which 825 are leaf classes) and 51,670 samples.
For the DL and ILP models, we use the same 80/10/10 train/validation/test split.

\subsubsection{DL model}
\begin{figure}[tb]
    \centering
    \includegraphics[width=0.8\linewidth]{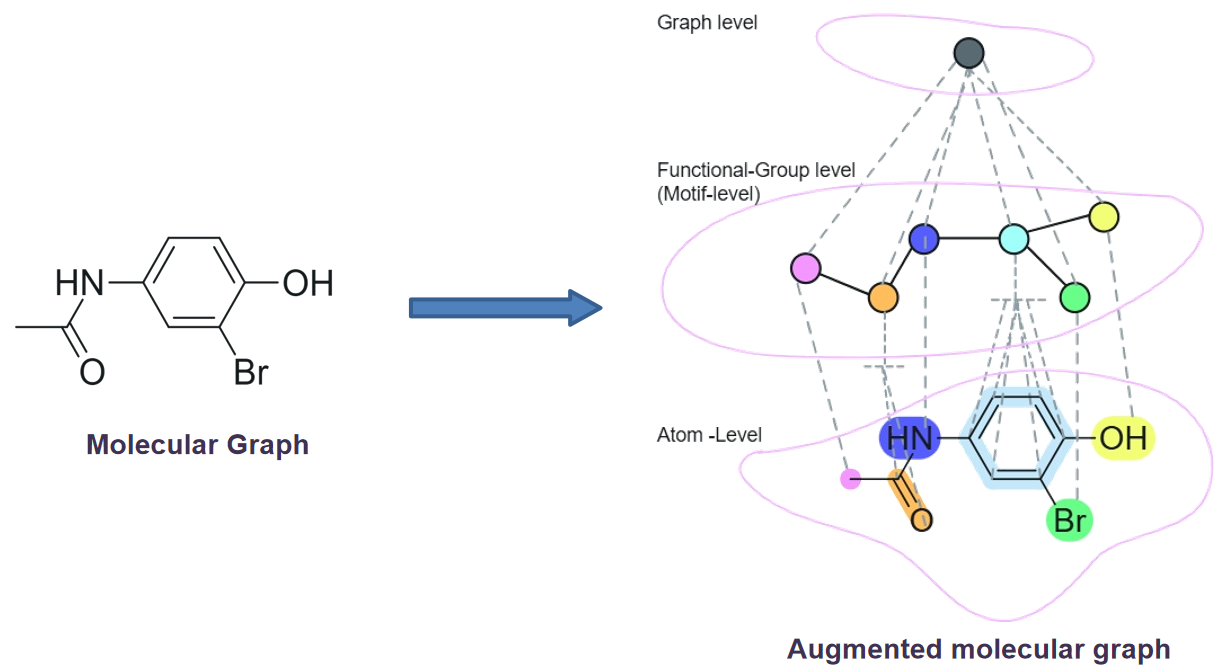}
    \caption{Representation of a molecule in the augmented graph used by the augmented GAT model. Figure taken from~\cite{khedekar2025integrating}, adapted.}
    \label{fig:augmented-molecule}
\end{figure}
As the DL model, we use the augmented Graph Attention Network (GAT) architecture developed in ~\cite{khedekar2025integrating}, which has been shown to outperform other Graph Neural Networks and Transformer models.
The augmented GAT consists of 3 connected levels, an atom level, a motif level and a global node (cf. Figure~\ref{fig:augmented-molecule}). 
Each level is initialised with different chemical properties (e.g., the atomic number and charge for the atom level nodes or the functional group name for motif level nodes).
The augmented GAT performs 4 message passing steps before aggregating all atom representations via sum pooling. 
The final classification is conducted by a fully connected classification head.

\subsubsection{ILP system}

\begin{table}[tb]
    \centering
    \caption{Predicates used to represent molecules as logic programs.}
    \begin{tabularx}{\textwidth} {  >{\centering\arraybackslash}X  >{\centering\arraybackslash}X }
        \hline
        Predicate & Description \\
        \hline
        \verb|chebi_{class_id}/1| & molecule belongs to ChEBI class \\
        \verb|has_atom/2| & atom belongs to molecule \\
        \hline
        \verb|c/1|, \verb|n/1|, ... & element symbols \\
        \verb|charge_p/1|, \verb|charge_n/1| & positive / negative atom charge \\
        \verb|charge0/1|, \verb|charge_m1/1|, \verb|charge1/1|, ... & specific atom charge \\
        \verb|has_0_hs/1|, \verb|has_1_hs/1|, ... & atom has this number of attached hydrogen atoms \\
        \verb|has_at_least_1_hs/1|, \verb|has_at_least_2_hs/1|, ... & atom has at least this number of attached hydrogen atoms \\
        \verb|cip_code_S/1|, \verb|cip_code_R/1| & stereoconfiguration according to CIP rules \\
        \verb|steroid_{n}| & atom is at steroid position n~\cite{IUPAC+S06005+2025} (up to 17) \\
        \hline
        \verb|has_bond_to/2| & two atoms share a covalent bond \\
        \verb|bSINGLE/2|, \verb|bDOUBLE/2|, ... & two atoms share a single, double, ... bond \\
        \hline
        \verb|net_charge_{positive|negative|neutral}| & global molecule charge \\
        \verb|aromatic/1|, \verb|aliphatic/1| & molecule is aromatic / aliphatic \\
        \hline
    \end{tabularx}
    \label{tab:lp-predicates}
\end{table}


\begin{figure}[tb]
    \centering
    \includegraphics[width=0.8\linewidth]{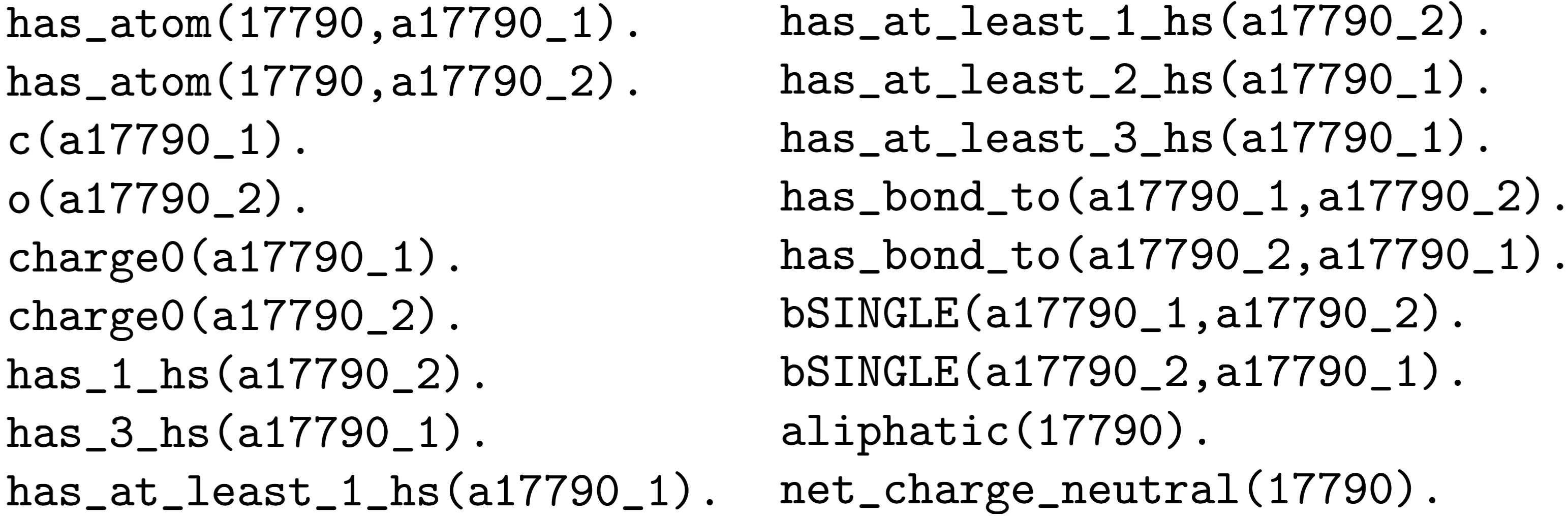}
    \caption{Representation of methanol (CH$_3$OH) as a logic program.}
    \label{fig:chebi-17790-bk}
\end{figure}
To formulate molecule classification as an ILP problem, we need to translate relevant molecule information into a logic program.
Table~\ref{tab:lp-predicates} shows the predicates used to represent different properties specific to bonds, atoms or molecules. 
An example for methanol, a molecule consisting of a carbon atom bonded to an oxygen atom, is shown in Figure~\ref{fig:chebi-17790-bk}. \verb|17790| represents the molecule itself, \verb|a17790_1| and \verb|a17790_2| represent the carbon and oxygen atoms.

Based on these molecule representations, an ILP problem is constructed for each label class. For performance reasons, not all molecules in the ChEBI25-3STAR training set are used for each class. Instead, 200 positive and negative samples are selected. For positive samples, the selection is random. For negative samples, a breadth first search is conducted on the label taxonomy: First, samples that belong to a sibling class of the target class are chosen. For instance, the negative samples for \textit{primary alcohol} are samples that belong to its direct superclass, \textit{alcohol} (CHEBI:30879), but not to \textit{primary alcohol} (e.g., secondary alcohols like \textit{propan-2-ol}, CHEBI:17824). If there are less than 25 such samples, samples that are one step further distant in the ChEBI graph are added (e.g., samples that are not alcohols, but belong to \textit{organic hydroxy compound}, CHEBI:33822). This is repeated until the minimum threshold of 25 is met after including all samples up to a given distance or the maximum of 200 samples is reached (in this case, the search is stopped immediately).
This method ensures that there is a sufficient number of negative samples for each target label and negative samples are as close as possible to the true samples.

The task of the ILP system is to find a rule that differentiates the positive from the negative samples.
For instance,

\begin{small}
\begin{verbatim}
  chebi_15734(M) :- has_atom(M,A1), has_atom(M,A2), bSINGLE(A1,A2), 
                      c(A1), has_at_least_2_hs(A1), o(A2).
\end{verbatim}
\end{small}

would be a good rule for \textit{primary alcohol}, since the statement ``methanol is a primary alcohol'', i.e., \verb|chebi_15734(17790).|, can be inferred from background knowledge (cf. Figure~\ref{fig:chebi-17790-bk}), but the statement ``propan-2-ol is a primary alcohol'' cannot be inferred (which would be false).

\subsection{Global explanations}
\label{sec:meth-global}

While the rules learned with ILP are symbolic, they are not automatically understandable for domain experts. 
To improve readability, we provide an algorithmic 
translation from ILP to Natural Language. 
The translation combines the learned ILP rule with the implicit superclass condition, retrieves class names from ChEBI and replaces predicate names with natural language expressions.

For instance, the rule for \textit{pyrazoles} (CHEBI:26410) is
\begin{small}
\begin{verbatim}
  chebi_26410(V0):- has_atom(V0,V1),bAROMATIC(V1,V2),n(V2),n(V1).
\end{verbatim}
\end{small}
It then gets translated as 
\begin{small}
\begin{verbatim}
  A compound is a pyrazoles (CHEBI:26410) if it
    (a) is a diazole (CHEBI:23677) AND
    (b) it has atoms A1 and A2. A1 is a nitrogen atom. 
        A1 has an aromatic bond to A2. A2 is a nitrogen atom.
\end{verbatim}  
\end{small}
If necessary, the literals of a clause get reordered such that all literals concerning a variable are as close to each other as possible.
Rules that consist of several clauses get translated as an ``or'' condition:
\begin{small}
\begin{verbatim}
  chebi_33296(V0) :- has_atom(V0,V1), li(V1).
  chebi_33296(V0) :- has_atom(V0,V1), na(V1).
  chebi_33296(V0) :- has_atom(V0,V1), k(V1).
\end{verbatim}
\end{small}
gets translated as 
\begin{small}
\begin{verbatim}
  A compound is an alkali metal molecular entity (CHEBI:33296) if it
  (a) is an s-block molecular entity (CHEBI:33674) AND
  (b) satisfies one of the following conditions:
    - it has an atom A1. A1 is a lithium atom. OR
    - it has an atom A2. A2 is a sodium atom. OR
    - it has an atom A3. A3 is a potassium atom.
\end{verbatim}
\end{small}

This approach allows us to discuss the ILP rules with experts for verification and potentially improvement.

\subsection{Local explanations}
\label{sec:meth-local}
A more intuitive approach to understanding the ILP rules is by example. 
Therefore, we also offer local explanations. 
Here, given a molecule, it gets traced which variables get grounded with which atoms using xclingo\footnote{https://github.com/bramucas/xclingo2}~\cite{Cabalar_2020}.

\begin{figure}[tb]
    \centering
    \includegraphics[width=1\linewidth]{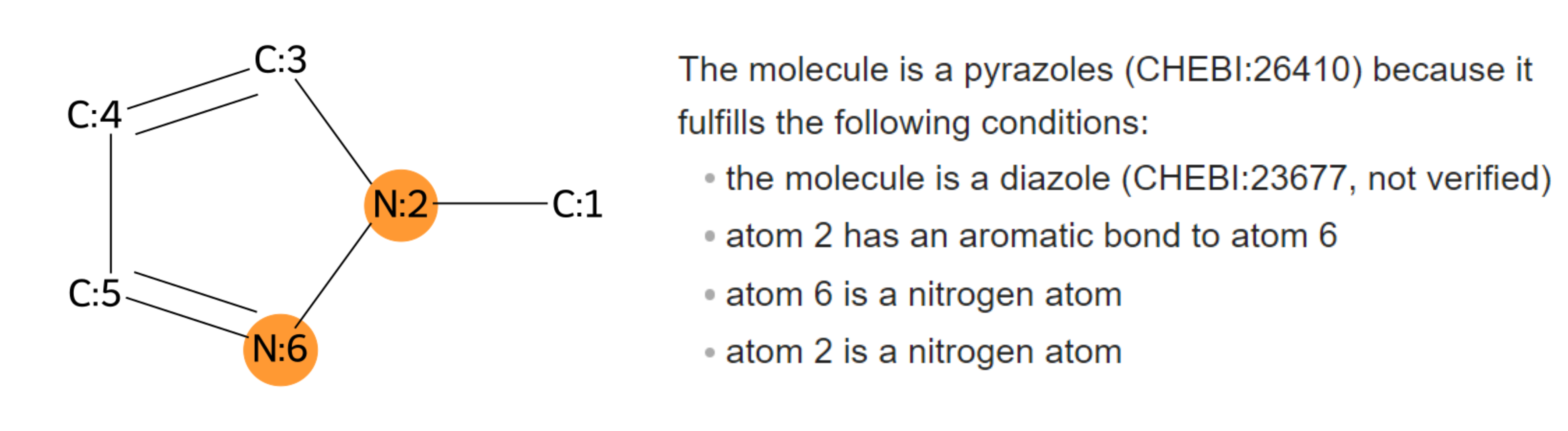}
    \caption{Local explanation for \textit{N-methylpyrazole}. It shows the molecular graph with numbered atoms and a textual description of the classification. Atoms relevant for the classification as \textit{pyrazoles} are highlighted.}
    \label{fig:chebi-59025}
\end{figure}
This is used to relate the global explanation to atoms in the molecular graph. For instance, for \textit{pyrazoles} (cf. Section~\ref{sec:meth-global}) and the molecule \textit{N-methylpyrazole} (CHEBI:59025), the explanation shown in Figure~\ref{fig:chebi-59025} is produced.

\subsection{Experimental configuration}
\label{sec:meth-exp}

The augmented GAT model has been trained on the ChEBI25-3STAR for 200 epochs with a learning rate of $10^{-3}$ and binary cross-entropy loss. The batch size was 512. Training has been conducted on a NVIDIA A100 GPU with the ChEB-AI Graph library\footnote{\url{https://github.com/ChEB-AI/python-chebai-graph}}. 

As mentioned in Section~\ref{sec:meth-ilp}, we use the Popper ILP system to learn ILP rules. 
The parametrised cost function is implemented in a forked version of Popper\footnote{\url{https://github.com/sfluegel05/Popper}}. For each cost function configuration (ILP-FP / ILP-FN / ILP-REG) and leaf class in the ChEBI25-3STAR dataset, we learn an ILP rule with a timeout of 60 seconds. Popper is used with the NuWLS anytime solver~\cite{chu2023nuwls} enabled and restrictions to a maximum number of 6 variables, 2 clauses and 8 body literals per rule. 

\section{Results}
\label{sec:results}
In this section, the HyDI ensemble is applied on the ChEBI25-3STAR dataset. 
Section~\ref{sec:res-ilp} evaluates the individual ILP rules for each configuration. In Section~\ref{sec:res-ensemble}, the ensemble as a whole is compared to the DL model baseline.

As the main evaluation metric, the F1 score (cf. Equation~\ref{eq:f1-score}) will be used. It can also be expressed as the harmonic mean between recall and precision, defined as 
\begin{equation}
    Precision(c) = \frac{\mathit{TP}(c)}{\mathit{TP}(c) + \mathit{FP}(c)} \quad \text{and} \quad Recall(c) = \frac{\mathit{TP}(c)}{\mathit{TP}(c) + \mathit{FN}(c)}
\end{equation}
for a class $c$. For a set of classes, these metrics can be aggregated either by taking the average over the class-wise metrics (macro-aggregation) or by summing up TPs, FPs and FNs for all classes (micro-aggregation).
All results in this Section are based on the ChEBI25-3STAR test set.

\subsection{ILP rule analysis}
\label{sec:res-ilp}
All three ILP configurations (ILP-FN, ILP-FP and ILP-REG) have been run on the 825 leaf classes. 
In some cases, Popper has not been able to produce a rule within the timeout of 60 seconds. 
For 149 classes, no configuration has produced a rule, while for 168 classes, one or two configurations have produced a rule.
\begin{table}[htbp]
\centering
\caption{Performance of all ILP-generated rules (independent of the ensemble). Shared classes are the 508 classes for which each ILP configuration has produced a rule. All metrics are macro-aggregated.}
\begin{tabular}{crrrrrrrr}
\hline
 & & \multicolumn{3}{c}{All classes with rules} & \multicolumn{3}{c}{Shared classes} \\
ILP configuration & \#rules & Precision & Recall & F1 score & Precision & Recall & F1 score \\
\hline
ILP-FN & 666 & 0.677 & \textbf{0.809} & 0.680 & 0.712 & \textbf{0.877} & \textbf{0.730} \\
ILP-FP & 520 & \textbf{0.732} & 0.666 & 0.640 & \textbf{0.734} & 0.667 & 0.642 \\
ILP-REG & 636 & 0.699 & 0.774 & \textbf{0.681} & 0.731 & 0.814 & 0.716 \\

\hline
\end{tabular}

\label{tab:ilp-results}
\end{table}

ILP-FN is the most reliable configuration with 666 produced rules (cf. Table~\ref{tab:ilp-results}) while ILP-FP has failed more often, only generating rules for 520 classes. In terms of performance, ILP-FP also falls short, with a macro F1 score 4\% / 4.1\% lower than ILP-FN / ILP-REG. When taking only the classes into account for which all configurations have produced rules, this gap widens to 8.8\% / 7.4\%. 
The precision and recall values explain where this gap comes from. 
While the ILP-FP configuration has the highest precision, its recall is significantly lower (by 21\% on the shared classes) than that of ILP-FN.

This can be explained by the different cost functions used: ILP-FN has a stronger weight for FNs, meaning that the ILP system is encouraged to find a rule with fewer FNs, while potentially allowing more FPs. This naturally translates to a better recall and lower precision. For ILP-FP the inverse happens. The cost function weights FPs stronger, bringing forth rules with high precision, but lower recall. 
Moreover, since ILP-FN produced more rules overall, the ``broader'' more inclusive rules targeted by ILP-FN seem to be easier to find by the ILP system than the ``narrower'', more eclectic rules for which ILP-FP aims. 
We hypothesise that broad rules are less complex and require less specific chemical knowledge. This makes them easier to identify within the constraints of the ILP system.

To illustrate the difference between ILP-FN and ILP-FP, take the class \textit{6$\beta$-hydroxy steroid} (CHEBI:36851). It is defined in ChEBI as ``Any 6-hydroxy steroid in which the 6-hydroxy substituent has $\beta$-configuration.''. 
A perfect ILP rule for this class would identify position 6 in the steroid structure and restrict the atom at this position to a $\beta$-configuration. In this context, the $\alpha$/$\beta$ refer to the spatial configuration of an atom. However, there is no predicate that expresses the $\alpha$ or $\beta$ configuration. Instead, the only available predicates for stereochemistry are the CIP codes R and S. While these do not consistently correspond to the $\alpha$ / $\beta$ configurations, in our dataset, most 6$\beta$ steroids have the CIP code R. Thus, we accept the following as a good rule for \textit{6$\beta$-hydroxy steroid}:
\begin{small}
\begin{verbatim}
    chebi_36851(V0) :- cip_code_R(V2), has_atom(V0,V2), steroid_6(V2).
\end{verbatim}
\end{small}
Note that the hydroxy group is not specified in this rule since it is already present in the superclass, \textit{6-hydroxy steroid} (CHEBI:36849). 
ILP-FN and ILP-FP both fail to find this rule, but for different reasons. With ILP-FN, one gets
\begin{small}
\begin{verbatim}
    chebi_36851(V0) :- has_1_hs(V1), has_atom(V0,V1), steroid_6(V1).
\end{verbatim}
\end{small}
This rule correctly identifies steroid position 6 and marks it as having 1 hydrogen atom (a side effect of the attached hydrogen group). However, it misses the stereochemical configuration, making it too general. On the test set, it receives a recall of 1, but a precision of only 0.6.
ILP-FP produces the rule
\begin{small}
\begin{verbatim}
    chebi_36851(V0) :- bSINGLE(V1,V2), cip_code_R(V2), has_2_hs(V1), 
                       has_atom(V0,V1), steroid_6(V2).
\end{verbatim}
\end{small}
Here, the steroid position and stereochemistry are present. But the rule also includes the condition that a neighbouring atom has to have 2 hydrogen atoms. Thus, it is overly specific. This rule gets a precision of 1, but a recall of 0.67.

Although ILP-FP get a lower overall F1 score, we have reason to assume that it can still contribute to the ensemble: Out of the 508 classes for which all configurations produced rules, ILP-FP gets a higher F1 score than ILP-REG and ILP-FN for 51 classes.

\subsection{Ensemble performance}
\label{sec:res-ensemble}
\begin{table}[tb]
    \centering
    \caption{Performance of the DL model vs. the HyDI ensemble on the ChEBI25-3STAR dataset.}
    \begin{tabular}{ccccccc}
    \hline
         & \multicolumn{3}{c}{Micro-aggregated} & \multicolumn{3}{c}{Macro-aggregated} \\
        Model & Precision & Recall & F1 score & Precision & Recall & F1 score \\
        \hline
        DL & \textbf{0.929} & \textbf{0.910} & \textbf{0.920} & \textbf{0.784} & \textbf{0.734} & \textbf{0.737} \\
        HyDI & 0.919 & \textbf{0.910} & 0.914 & 0.782 & 0.729 & 0.731 \\
        \hline
    \end{tabular}
    \label{tab:dl-hydi-performance}
\end{table}
Next, we evaluate the ensemble as a whole. As the baseline, we use the augmented GAT model that is also used as the DL component of the ensemble.
In total, 314 of the leaf classes are predicted by an ILP rule and 511 are predicted by the DL model.
Table~\ref{tab:dl-hydi-performance} shows the performance of the HyDI ensemble and the DL baseline across all classes. 
The DL model on its own performs slightly better than the ensemble (0.6\% for both the micro / macro F1 score).
For the micro-F1 score, this difference comes mostly from the lower precision, while the recall stays the same. 
With macro-aggregation, the precision of DL and HyDI is nearly the same while the recall drops slightly.

\begin{table}[tb]
    \centering
    \caption{Performance for the ChEBI subsets predicted by each ILP type / DL, macro-aggregated.}
    \begin{tabular}{crcccccc}
    \hline
         &  & \multicolumn{2}{c}{Precision} & \multicolumn{2}{c}{Recall} & \multicolumn{2}{c}{F1 score} \\
        Class subset & \# classes & DL & HyDI & DL & HyDI & DL & HyDI \\
        \hline
        DL & 1,494 & 0.794 & 0.794 & 0.741 & 0.741 & 0.746 & 0.746 \\
        \hline
ILP-FN & 260 & 0.730 & \textbf{0.735} & \textbf{0.712} & 0.700 & \textbf{0.698} & 0.680 \\
ILP-REG & 24 & \textbf{0.766} & 0.558 & \textbf{0.640} & 0.524 & \textbf{0.674} & 0.481 \\
ILP-FP & 30 & \textbf{0.738} & 0.732 & \textbf{0.664} & 0.602 & \textbf{0.668} & 0.613 \\
    \hline
    \end{tabular}
    \label{tab:dl-hydi-by-subset}
\end{table}

Table~\ref{tab:dl-hydi-by-subset} breaks down the prediction scores by subsets which are either predicted by the DL model or by one of the ILP configurations. 
Most of the selected rules (260 out of 314) are from the ILP-FN configuration. 
However, this does not mean that ILP-FN produced a better rule for all of these classes. 
For classes where several rules receive the same validation F1 score, the ILP-FN rule was chosen based on the configuration's macro F1 score.
A noticeable outlier is the ILP-REG subset. Here, the ILP-generated rules perform significantly worse than on other subsets (-19.9\% / -13.2\% compared to ILP-FN / ILP-FP) and the DL model on the ILP-REG subset (-19.3\%).
Apparently, the validation set performance does not generalise towards the test set for this subset.

This has several reasons. For one, the test and validation sets, especially for leaf classes, can be relatively small. 
Since the sets each get 10\% of, in the worst case, 25 samples, some classes only have 2 or 3 positive samples for validation / testing. Thus, differences in the F1 score of 50\% or more between the validation and test set are statistically plausible. 
Taking a closer look at the classes for which ILP-REG rules were selected, it becomes clear why not all of them can be learned adequately. Classes like \textit{bioconjugate} (CHEBI:64985) are not structurally defined. 
Other classes can be described by structural features, but not with the chosen ILP problem setting. For instance, \textit{diterpene} (CHEBI:35190) is a terpene with 20 carbon atoms. 
Counting the exact number of carbon atoms in a molecule is not feasible with the current approach.

\section{Conclusions}
\label{sec:conclusions}
In this work, we have presented the HyDI ensemble, a methodology for combining Deep Learning models with ILP rules to get more explainable classification results in hierarchical multi-label classification (HMLC) tasks. 
While the upper layers of the hierarchy get predicted by a Deep Learning multi-label classifier, for the leaf classes, individual ILP rules get learned.
The ensemble methodology has been applied to a real-world HMLC problem from chemistry based on the ChEBI ontology. For 314 out of 825 leaf classes, the Deep Learning model has been replaced by an ILP rule. 
While some of the learned rules are chemically meaningful, others show signs of overfitting or try to capture concepts that cannot be formalised with the available predicates. Altogether, the evaluation shows that this only leads to a minimal loss in performance (0.6\% in the macro F1 score). Simultaneously, interpretability is greatly enhanced through ILP rules that can be translated into natural language and visualised in terms of the input molecule.

In future work, we plan to improve the ILP rules with pre-defined functional groups and ring structures. This will allow the ILP system to capture more complex chemical concepts. Beyond that, we will use predicate invention and negation to enhance the ILP system's capabilities.
Another research avenue will be to iteratively refine rules with human experts. Here, the ILP rules will be a starting point. 
The global and local explanations proposed in this work will help experts to identify issues and make changes to the rules which then can get automatically verified.

\begin{credits}
\subsubsection{\ackname} This study was funded
by the Deutsche Forschungsgemeinschaft (DFG, German Research Foundation) - 522907718 and 456666331.

We would like to thank Adnan Malik (Chemical Biology Services, EMBL's European Bioinformatics Institute), Kseniya Zubko (Faculty of Computer Science, Otto-von-Guericke University Magdeburg) and Prof. Edgar Haak for helping us with their chemical expertise in the evaluation.
\subsubsection{\discintname}
The authors have no competing interests to declare that are
relevant to the content of this article. 
\end{credits}

\subsection*{Supplementary Material}
Supplementary material is available on Zenodo\footnote{\url{https://zenodo.org/records/20411715}}.
The code for creating the dataset, running the ILP system and building the ensemble is available on GitHub\footnote{\url{https://github.com/ChEB-AI/chebILP}}.
The DL model has been trained with the ChEB-AI graph library\footnote{\url{https://github.com/ChEB-AI/python-chebai-graph}}\footnote{Model weights: \url{https://huggingface.co/chebai/gat-aug-chebi25-3star\_v248}}.

The ChEBI25-3STAR dataset and the ILP problems created from this dataset are available on Hugging Face\footnote{\url{https://huggingface.co/datasets/chebai/ChEBI25-3STAR}}\footnote{\url{https://huggingface.co/datasets/chebai/ChEBI25-3STAR-ILP}}.

%
%
%
\bibliographystyle{splncs04}
\bibliography{bibliography}

@misc{IUPAC+S06005+2025,
institution = {International Union of Pure and Applied Chemistry (IUPAC)},
year = {2025},
title = {steroids},
doi = {doi:10.1351/goldbook.S06005},
edition = {5.0.0},
author = {{IUPAC}}
}

@article{zhang2023critical,
  title={A critical review of inductive logic programming techniques for explainable AI},
  author={Zhang, Zheng and Yilmaz, Levent and Liu, Bo},
  journal={IEEE transactions on neural networks and learning systems},
  volume={35},
  number={8},
  pages={10220--10236},
  year={2023},
  publisher={IEEE}
}

@article{muggleton1991inductive,
  title={Inductive logic programming},
  author={Muggleton, Stephen},
  journal={New generation computing},
  volume={8},
  number={4},
  pages={295--318},
  year={1991},
  publisher={Springer}
}

@article{cropperInductiveLogicProgramming2022,
  title = {Inductive {{Logic Programming At}} 30: {{A New Introduction}}},
  shorttitle = {Inductive {{Logic Programming At}} 30},
  author = {Cropper, Andrew and Duman{\v c}i{\'c}, Sebastijan},
  year = 2022,
  journal = {Journal of Artificial Intelligence Research},
  volume = {74},
  pages = {765--850},
  issn = {1076-9757},
  doi = {10.1613/jair.1.13507},
  urldate = {2026-05-20},
  copyright = {Copyright (c)},
  langid = {english}
}

@article{cropperLearningProgramsLearning2021,
  title = {Learning Programs by Learning from Failures},
  author = {Cropper, Andrew and Morel, Rolf},
  year = 2021,
  journal = {Machine Learning},
  volume = {110},
  number = {4},
  pages = {801--856},
  issn = {1573-0565},
  doi = {10.1007/s10994-020-05934-z},
  urldate = {2025-10-06},
  langid = {english}
}

@article{cropperLearningLogicPrograms2023,
  title = {Learning {{Logic Programs}} by {{Discovering Where Not}} to {{Search}}},
  author = {Cropper, Andrew and Hocquette, C{\'e}line},
  year = 2023,
  journal = {Proceedings of the AAAI Conference on Artificial Intelligence},
  volume = {37},
  number = {5},
  pages = {6289--6296},
  issn = {2374-3468},
  doi = {10.1609/aaai.v37i5.25774},
  urldate = {2026-05-20},
  copyright = {Copyright (c) 2023 Association for the Advancement of Artificial Intelligence},
  langid = {english}
}

@inproceedings{raboldEnrichingVisualVerbal2020,
  title = {Enriching {{Visual}} with {{Verbal Explanations}} for {{Relational Concepts}} -- {{Combining LIME}} with {{Aleph}}},
  booktitle = {Machine {{Learning}} and {{Knowledge Discovery}} in {{Databases}}},
  author = {Rabold, Johannes and Deininger, Hannah and Siebers, Michael and Schmid, Ute},
  editor = {Cellier, Peggy and Driessens, Kurt},
  year = 2020,
  pages = {180--192},
  publisher = {Springer International Publishing},
  address = {Cham},
  doi = {10.1007/978-3-030-43823-4_16},
  isbn = {978-3-030-43823-4},
  langid = {english}
}

@inproceedings{raboldExplainingBlackBoxClassifiers2018,
  title = {Explaining {{Black-Box Classifiers}} with {{ILP}} -- {{Empowering LIME}} with {{Aleph}} to {{Approximate Non-linear Decisions}} with {{Relational Rules}}},
  booktitle = {Inductive {{Logic Programming}}},
  author = {Rabold, Johannes and Siebers, Michael and Schmid, Ute},
  editor = {Riguzzi, Fabrizio and Bellodi, Elena and Zese, Riccardo},
  year = 2018,
  pages = {105--117},
  publisher = {Springer International Publishing},
  address = {Cham},
  doi = {10.1007/978-3-319-99960-9_7},
  isbn = {978-3-319-99960-9},
  langid = {english}
}

@inproceedings{raboldExpressiveExplanationsDNNs2020,
  title = {Expressive {{Explanations}} of {{DNNs}} by~{{Combining Concept Analysis}} with {{ILP}}},
  booktitle = {{{KI}} 2020: {{Advances}} in {{Artificial Intelligence}}},
  author = {Rabold, Johannes and Schwalbe, Gesina and Schmid, Ute},
  editor = {Schmid, Ute and Kl{\"u}gl, Franziska and Wolter, Diedrich},
  year = 2020,
  pages = {148--162},
  publisher = {Springer International Publishing},
  address = {Cham},
  doi = {10.1007/978-3-030-58285-2_11},
  isbn = {978-3-030-58285-2},
  langid = {english}
}

@inproceedings{ribeiro2016should,
  title={"Why should {{I}} trust you?" {{E}}xplaining the predictions of any classifier},
  author={Ribeiro, Marco Tulio and Singh, Sameer and Guestrin, Carlos},
  booktitle={Proceedings of the 22nd ACM SIGKDD international conference on knowledge discovery and data mining},
  pages={1135--1144},
  year={2016}
}

@misc{srinivasan2001aleph,
  title={The {{Aleph}} manual},
  author={Srinivasan, Ashwin},
  howpublished = {\url{https://www.cs.ox.ac.uk/activities/programinduction/Aleph/aleph.html}},
  note = {Accessed: 2026-06-05},
  year={2001}
}

@article{srinivasan1996theories,
  title={Theories for mutagenicity: A study in first-order and feature-based induction},
  author={Srinivasan, Ashwin and Muggleton, Stephen H and Sternberg, Michael JE and King, Ross D},
  journal={Artificial Intelligence},
  volume={85},
  number={1-2},
  pages={277--299},
  year={1996},
  publisher={Elsevier}
}

@inproceedings{dasaroInductiveLogicProgramming2020,
  title = {Towards an {{Inductive Logic Programming Approach}} for {{Explaining Black-Box Preference Learning Systems}}},
  booktitle = {Proceedings of the {{Seventeenth International Conference}} on {{Principles}} of {{Knowledge Representation}} and {{Reasoning}}},
  author = {D'Asaro, Fabio A. and Spezialetti, Matteo and Raggioli, Luca and Rossi, Silvia},
  year = 2020,
  month = jul,
  pages = {855--859},
  publisher = {International Joint Conferences on Artificial Intelligence Organization},
  address = {Rhodes, Greece},
  doi = {10.24963/kr.2020/88},
  urldate = {2026-05-20},
  isbn = {978-0-9992411-7-2},
  langid = {english}
}

@article{law2020ilasp,
  title={The {{ILASP}} system for inductive learning of answer set programs},
  author={Law, Mark and Russo, Alessandra and Broda, Krysia},
  journal={arXiv preprint arXiv:2005.00904},
  year={2020}
}

@article{glauer2024interpretable,
  title={Interpretable ontology extension in chemistry},
  author={Glauer, Martin and Memariani, Adel and Neuhaus, Fabian and Mossakowski, Till and Hastings, Janna},
  journal={Semantic Web},
  volume={15},
  number={4},
  pages={937--958},
  year={2024},
  publisher={SAGE Publications Sage UK: London, England}
}

@article{glauer2024chebifier,
  title={{{Chebifier}}: automating semantic classification in {{ChEBI}} to accelerate data-driven discovery},
  author={Glauer, Martin and Neuhaus, Fabian and Fl{\"u}gel, Simon and Wosny, Marie and Mossakowski, Till and Memariani, Adel and Schwerdt, Johannes and Hastings, Janna},
  journal={Digital Discovery},
  volume={3},
  number={5},
  pages={896--907},
  year={2024},
  publisher={Royal Society of Chemistry}
}

@article{memariani2025box,
  title={Box embeddings for extending ontologies: a data-driven and interpretable approach},
  author={Memariani, Adel and Glauer, Martin and Fl{\"u}gel, Simon and Neuhaus, Fabian and Hastings, Janna and Mossakowski, Till},
  journal={Journal of Cheminformatics},
  volume={17},
  number={1},
  pages={138},
  year={2025},
  publisher={Springer}
}

@article{khedekar2025integrating,
  title={Integrating Chemical Knowledge into Graph Neural Networks},
  author={Khedekar, Aditya Ganesh},
  journal={Master's Thesis},
  year={2025}
}

@misc{flugelHeterogeneousEnsembleLearning2026,
  title = {Heterogeneous {{Ensemble Learning}} for {{Hierarchical Multi-label Classification}}},
  author = {Fl{\"u}gel, Simon and Glauer, Martin and Hastings, Janna and Mossakowski, Till and Mungall, Christopher J. and Tumescheit, Charlotte and Neuhaus, Fabian and Khedekar, Aditya Ganesh},
  year = 2026,
  month = may,
  publisher = {Research Square},
  issn = {2693-5015},
  doi = {10.21203/rs.3.rs-9023090/v1},
  urldate = {2026-06-03},
  archiveprefix = {Research Square}
}

@article{magkaRulebasedOntologicalFramework2014,
  title = {A Rule-Based Ontological Framework for the Classification of Molecules},
  author = {Magka, Despoina and Kr{\"o}tzsch, Markus and Horrocks, Ian},
  year = 2014,
  month = apr,
  journal = {Journal of Biomedical Semantics},
  volume = {5},
  number = {1},
  pages = {17},
  issn = {2041-1480},
  doi = {10.1186/2041-1480-5-17},
  urldate = {2025-12-04},
  langid = {english}
}

@article{flugel2025chemlog,
  title={ChemLog: Making MSOL viable for ontological classification and learning},
  author={Fl{\"u}gel, Simon and Glauer, Martin and Mossakowski, Till and Neuhaus, Fabian},
  journal={arXiv preprint arXiv:2507.13987},
  year={2025}
}

@article{mungall2025chemical,
  title={Chemical classification program synthesis using generative artificial intelligence},
  author={Mungall, Christopher J and Malik, Adnan and Korn, Daniel R and Reese, Justin T and O’Boyle, Noel M and Hastings, Janna},
  journal={Journal of Cheminformatics},
  volume={17},
  number={1},
  pages={152},
  year={2025},
  publisher={Springer}
}

@article{flugel2026defining,
  title={Defining peptides in {{ChEBI}}},
  author={Fl{\"u}gel, Simon and Mossakowski, Till and Neuhaus, Fabian and Pfanenstiel, Erik and Glauer, Martin and Haak, Edgar and Malik, Adnan and O’Boyle, Noel M},
  journal={Journal of Cheminformatics},
  year={2026},
  publisher={Springer}
}

@article{malik2026chebi,
  title={{{ChEBI}}: re-engineered for a sustainable future},
  author={Malik, Adnan and Arsalan, Muhammad and Moreno, Carlos and Mosquera, Juan and F{\'e}lix, Eloy and Kizil{\"o}ren, Tevfik and Muthukrishnan, Venkatesh and Zdrazil, Barbara and Leach, Andrew R and O’Boyle, Noel M},
  journal={Nucleic Acids Research},
  volume={54},
  number={D1},
  pages={D1768--D1778},
  year={2026},
  publisher={Oxford University Press}
}

@inproceedings{chu2023nuwls,
  title={{{NuWLS}}: Improving local search for (weighted) partial {{MaxSAT}} by new weighting techniques},
  author={Chu, Yi and Cai, Shaowei and Luo, Chuan},
  booktitle={Proceedings of the AAAI Conference on Artificial Intelligence},
  volume={37},
  pages={3915--3923},
  year={2023}
}

@article{Cabalar_2020,
   title={A System for Explainable Answer Set Programming},
   volume={325},
   ISSN={2075-2180},
   url={http://dx.doi.org/10.4204/EPTCS.325.19},
   DOI={10.4204/eptcs.325.19},
   journal={Electronic Proceedings in Theoretical Computer Science},
   publisher={Open Publishing Association},
   author={Cabalar, Pedro and Fandinno, Jorge and Muñiz, Brais},
   year={2020},
   month=Sept, pages={124–136} }
\end{document}